\documentclass[conference]{IEEEtran}
\IEEEoverridecommandlockouts
\usepackage{cite}
\usepackage{amsmath,amssymb,amsfonts}
\usepackage{algorithmic}
\usepackage{graphicx}
\usepackage{textcomp}
\usepackage{xcolor}
\usepackage{enumitem}
\usepackage{amsmath}
\usepackage{amsthm}
\usepackage{amsfonts}
\usepackage{xcolor}

\theoremstyle{definition}
\newtheorem{definition}{Definition}
\def\BibTeX{{\rm B\kern-.05em{\sc i\kern-.025em b}\kern-.08em
    T\kern-.1667em\lower.7ex\hbox{E}\kern-.125emX}}

\begin{document}
\title{A Survey of Generalization of Graph Anomaly Detection: From Transfer Learning to Foundation Models\\
\thanks{ \textsuperscript{†}Corresponding author. \\
This research was partly funded by the Australian Research Council (ARC) under grant DP240101547.}
}

\author{
\IEEEauthorblockN{Junjun Pan}
\IEEEauthorblockA{\textit{School of ICT} \\
\textit{Griffith University}\\
Gold Coast, Australia \\
junjun.pan@griffithuni.edu.au}
\and
\IEEEauthorblockN{Yu Zheng}
\IEEEauthorblockA{\textit{School of ICT} \\
\textit{Griffith University}\\
Gold Coast, Australia \\
yu.zheng@griffith.edu.au}
\and
\IEEEauthorblockN{Yue Tan}
\IEEEauthorblockA{\textit{School of CSE} \\
\textit{University of New South Wales}\\
Sydney, Australia \\
yue.tan@unsw.edu.au}
\and
\IEEEauthorblockN{Yixin Liu\textsuperscript{†}}
\IEEEauthorblockA{\textit{School of ICT} \\
\textit{Griffith University}\\
Gold Coast, Australia \\
yixin.liu@griffith.edu.au}
}

\maketitle

\begin{abstract}
Graph anomaly detection (GAD) has attracted increasing attention in recent years for identifying malicious samples in a wide range of graph-based applications, such as social media and e-commerce. However, most GAD methods assume identical training and testing distributions and are tailored to specific tasks, resulting in limited adaptability to real-world scenarios such as shifting data distributions and scarce training samples in new applications. To address the limitations, recent work has focused on improving the generalization capability of GAD models through \textit{transfer learning} that leverages knowledge from related domains to enhance detection performance, or developing ``one-for-all'' GAD \textit{foundation models} that generalize across multiple applications. Since a systematic understanding of generalization in GAD is still lacking, in this paper, we provide a comprehensive review of generalization in GAD. We first trace the evolution of generalization in GAD and formalize the problem settings, which further leads to our systematic taxonomy. Rooted in this fine-grained taxonomy, an up-to-date and comprehensive review is conducted for the existing generalized GAD methods. Finally, we identify current open challenges and suggest future directions to inspire future research in this emerging field. 
\end{abstract}

\maketitle

\section{Introduction}

With the advances in information technology, graph-structured data has become a ubiquitous data structure in online services, including social media~\cite{li2021relevance}, e-commerce~\cite{xu2019relation}, and autonomous agents~\cite{liu2025graph,li2025assemble,shen2025understanding,miao2025blindguard}. This widespread usage has led to a significant increase in various malicious activities, including hacking, spam, and fake news. In order to identify these anomalous entities and behaviors from graph-structured data, graph anomaly detection (GAD) has emerged as an active research topic in recent years. To date, GAD has been applied to a wide range of domains, including but not limited to cybersecurity, financial fraud detection, recommender systems, and social network analysis, where identifying anomalous patterns is crucial for maintaining system integrity and user trust~\cite{qiao2024deep,liu2022bond, ma2021comprehensive,tang2023gadbench,liu2024self,ding2024divide}.

\begin{figure*}

\includegraphics[width=1.0\linewidth]{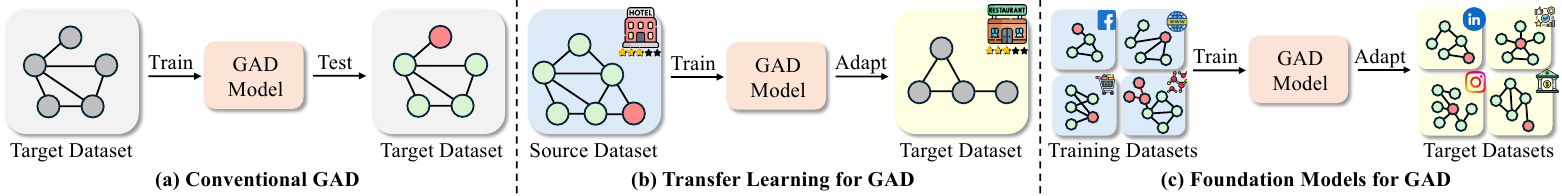}
\vspace{-6mm}
\caption{The learning paradigms of (a) conventional GAD methods; (b) transfer learning for GAD; and (c) foundation models for GAD.}
\label{fig:intro}
\vspace{-3mm}
\end{figure*}

Despite the growing popularity of GAD, conventional GAD paradigm (Fig.~\ref{fig:intro}(a)) usually operate under well-controlled \textit{in vitro} settings: On one hand, they typically assume that the training and testing sets are drawn from the same distribution, making them hard to transfer to other domains or unseen data; On the other hand, models are often tailored to a specific GAD task or scenario, limiting their generalizability across different application contexts. These assumptions limit the robustness and flexibility of current approaches in real-world settings, where GAD models are expected to adapt to varying data distributions and diverse scenarios~\cite{han2023anomaly, zhou2022domain}. For example, in the application of fraud detection, where transaction networks continuously evolve over time, the same-distribution assumption may hinder GAD models from adapting to emerging fraudulent patterns and distribution shifts. Moreover, in data-scarce or privacy-sensitive scenarios, such as healthcare and finance, training a specific GAD model can be challenging due to limited access to annotated data and strict privacy constraints~\cite{liu2024arc}.

In order to enhance the practicality of GAD and extend its applicability to more real-world scenarios, a recently emerging trend is to \textit{improve the generalization capability} of GAD methods. One promising research direction is to empower GAD methods with \textbf{transfer learning}~(Fig.~\ref{fig:intro}(b)), where knowledge from related source datasets is leveraged to improve anomaly detection on a similar target dataset. Using various techniques to learn transferable knowledge and capture target-domain anomaly patterns, recent transfer learning-based GAD methods can exploit data from related domains to build more powerful detection models, thereby enhancing generalization and reducing data dependency~\cite{ding2021cross,wang2023cross}. Following the success of large-scale pre-trained models~\cite{zhou2022domain}, a recent emerging direction is to build \textbf{foundation models}~(Fig.~\ref{fig:intro}(c)) for GAD, which are capable of generalizing across diverse anomaly detection scenarios in graph data. Under this direction, advanced approaches have achieved GAD at multiple granularities~\cite{lin2024unigad} as well as zero-shot GAD on arbitrary unseen datasets~\cite{niu2024zero}. 

Despite the growing research trend toward generalization of GAD, there is still no systematic review and categorization of studies in this field. This highlights the need for a comprehensive survey to organize existing work and guide future research. 
To fill the gap, in this paper, we provide a comprehensive and systematic survey of generalization in GAD. Specifically, the contributions of this paper are:

\begin{itemize}[leftmargin=*, labelsep=0.5em]
    \item \textbf{Problem Formulation}. We discuss the evolution of generalizability in GAD, highlighting the problem formulations and the underlying motivations.
    \item \textbf{Taxonomy}. Under the umbrella of two generalized paradigms, namely \textit{transfer learning} and \textit{foundation models}, we develop a taxonomy to organize existing generalized GAD approaches into more fine-grained categories.
    \item \textbf{Timely Review}. For each category, we offer a comprehensive review of recent advances, discussing the underlying motivations and design principles.
    \item \textbf{Future Directions}. We outline several open research directions to guide future research of this promising topic.
\end{itemize}

\section{Problem Formulation and Taxonomy}
In this section, we introduce the notations and problem statement in graph anomaly detection (GAD) and provide an overview of the taxonomy that illustrates increasing levels of generalizability. We begin with the statement of conventional GAD, then move to transfer learning approaches that enhance the generalizability of GAD methods within similar application scenarios. Finally, we present GAD foundation models that support broader generalization across different anomaly granularities and application domains.

\noindent\textbf{Notation.} 
An attributed graph is denoted as $\mathcal{G} = (\mathcal{V}, \mathcal{E}, \mathbf{X})$, where $\mathcal{V}$ and $\mathcal{E}$  are the node and edge sets respectively. $\mathbf{X} \in \mathbb{R}^{|\mathcal{V}| \times d}$ denotes the feature matrix of graph $\mathcal{G}$ with feature dimension $d$. A dataset $\mathcal{D}$ is defined as a single graph, i.e., $\mathcal{D}=\mathcal{G}$ or a homogeneous collection of graphs that share the same feature dimensionality and semantic space, i.e., $\mathcal{D}=\{\mathcal{G}_1, ... \mathcal{G}_N\}$. A \textit{sample} can be defined as the portion of graphs that is of interest for a specific task. For instance, in node-level tasks and edge-level tasks, each node and edge in the graphs serves as a sample, respectively; while in graph-level tasks, each entire graph is treated as a sample.

\subsection{Conventional Graph Anomaly Detection}

GAD can be defined as the task of identifying abnormal or rare samples in graphs that deviate significantly from expected structures or attributes~\cite{pan2025label,zhao2025freegad}. These samples can be nodes, edges, subgraphs (motifs), or entire graphs. %
\begin{definition}[Conventional GAD]
\label{def:conventional_gad} 
Given a graph dataset $\mathcal{D}$, GAD aims to learn an anomaly scoring function $f(\cdot)$ that assigns a score $s = f(o)$ to each sample $o$ in the dataset. Here $o$ can be a node, an edge, a subgraph, or a graphs, depending on the level of granularity considered. $f(\cdot)$ is expected to assign lower anomaly scores to normal samples and higher scores to the anomalous ones. Conventional GAD methods typically assume that the data for GAD model training and evaluation belong to the same dataset $\mathcal{D}$.
\end{definition}

\noindent\textbf{Limitation.} Despite the progress made by the conventional GAD methods, they are often tightly coupled to a specific training dataset or domain and exhibit poor generalization to unseen graph data~\cite{zhou2023improving}, particularly under distributional or domain shifts~\cite{gao2023alleviating, ding2021cross}. 
This insufficiency limits their robustness and flexibility in real-world scenarios~\cite{wu2024graph}. For example, anomaly patterns may evolve over time~\cite{gao2023alleviating, wang2024nsreg}, and training data are often scarce in the early stages of graph application deployment~\cite{han2023anomaly}. Even between the training and testing splits of a single graph, distribution shift can exist and degrade the GAD performance~\cite{gao2023alleviating}. Consequently, researchers have increasingly focused on enhancing the generalizability of GAD methods to better satisfy the demands of real-world applications.

\subsection{Transfer Learning for GAD}
To enhance generalizability across domains, researchers have explored transfer learning for GAD, which utilizes the knowledge learned from one graph dataset (the source domain) to enhance anomaly detection on another (the target domain). This capability is especially beneficial in guarding real-world graph applications: On one hand, it allows the model to leverage additional data or annotations from related domains when the target domain is limited in data; On the other hand, it improves the robustness of GAD methods to better adapt to evolving anomaly types. %

\begin{definition}[Transfer Learning for GAD]
\label{def:transferlearning_gad} 
The goal of transfer learning-based GAD models is to learn an anomaly scoring function $f(\cdot)$ that aims to identify anomalies in a target dataset $\mathcal{D}^t$ by utilizing additional data resources from one or more source datasets $\mathcal{D}^s$. To enable effective knowledge transfer, two key assumptions are typically made: 1) There exists common knowledge between $\mathcal{D}^t$ and $\mathcal{D}^s$, such as shared sample semantics and anomaly patterns. 2) The difference between $\mathcal{D}^t$ and $\mathcal{D}^s$ is moderate to allow the reuse or fine-tuning of GAD models across them. This requires alignment in feature dimensionality, semantic space, and anomaly types.

\end{definition}

\noindent\textbf{Taxonomy.} 
In this paper, we review methods of transfer learning for GAD following a problem-oriented taxonomy. Specifically, we identify two key challenges that arise from the underlying assumptions in transfer learning: 1) how to extract transferable knowledge across domains, and 2) how to capture target-specific patterns. Motivated by the first challenges, we summarize the techniques for \textit{\textbf{learning transferable knowledge}} in Section 3.1, which includes generalization-centric training and source-target representation alignment. Then, in Section 3.2, the GAD approaches for \textit{\textbf{capturing target-specific patterns}} are listed, including target-aware pre-training and test-time fine-tuning.

\noindent\textbf{Limitation.} 
While transfer learning-based methods have marked a key step toward generalizable GAD, their generalizability remains restricted due to the strong assumption of moderated domain discrepancy, which limits their applications in several data-scarce and privacy-sensitive real-world scenarios. In this case, more flexible models that can identify anomalies across different scenarios and domains are expected to expand the generalizability of GAD methods.

\subsection{Foundation Models for GAD}
To overcome the above limitations, GAD foundation models are an advanced solution by learning a one-for-all model for anomaly detection on various graphs in the wild, enabling generalization across a wide range of tasks and application scenarios. Compared to transfer learning, GAD foundation models offer stronger knowledge transferability and better scalability, and can even support zero-shot anomaly detection on previously unseen graphs~\cite{qiao2025anomalygfm}. Unlike conventional GAD or transfer learning approaches that are typically tailored to a specific anomaly pattern or application domain, GAD foundation models are built with inherent multi-task capabilities that support generalization across different detection settings.  

\begin{definition}[Foundation Models for GAD]
GAD foundation models are trained on either one dataset $\mathcal{D}^{tr}$ or a collection of training datasets $\mathcal{T}^{tr}=\{\mathcal{D}^{tr}_1, \cdots, \mathcal{D}^{tr}_n\}$, where $\mathcal{D}^{tr}_i$ is from an arbitrary domain. Ideally, a GAD foundation model is a scoring function $f(\cdot)$ that is able to predict an anomaly score for an arbitrary sample $o$, where $o$ can: 1) belong to any unseen dataset $\mathcal{D}^{te}_i \in \mathcal{T}^{te}$ in the wild that satisfies $\mathcal{D}^{te}_i \notin \mathcal{T}^{tr}$ and even does not originate from the same domain as any of the training datasets, and 2) be associated with different granularities, such as nodes, edges, subgraphs, and entire graphs.
\end{definition}

\noindent\textbf{Taxonomy.} 
As an emerging field, current GAD foundational models tend to focus their research on either cross-granularity generalization or cross-scenario (i.e., cross-dataset) generalization, naturally forming our taxonomy. Specifically, in Section 4.1, we introduce \textit{\textbf{cross-granularity GAD foundation models}} that can identify anomalies at multiple levels of graph granularity. Then, in Section 4.2, we summarize \textit{\textbf{cross-scenario GAD foundation models}} that are trained on diverse datasets and can predict on previously unseen datasets.

\noindent\textbf{Prospects.} 
Owing to their strong cross-granularity and cross-scenario generalization capability, GAD foundation models are considered a promising and forward-looking direction. Their flexibility and generalization potential make them especially suited to broad real-world applications.

\section{Transfer Learning for GAD}

Transfer learning-based GAD methods aim to leverage the knowledge from additional graphs in related domains (i.e., source data) to build a more powerful detection model on an application graph (i.e., target data) where data or labels may be scarce. To achieve positive transfer from source to target data, it is crucial to extract transferable knowledge from the source domains while capturing target-specific anomaly patterns during the learning process. 
Focusing on addressing each of these challenges, in this section, we review the representative GAD studies that aim to \textit{learn transferable knowledge} and \textit{capture target-specific patterns}.

\begin{figure}
\centering
\includegraphics[width=0.9\linewidth]{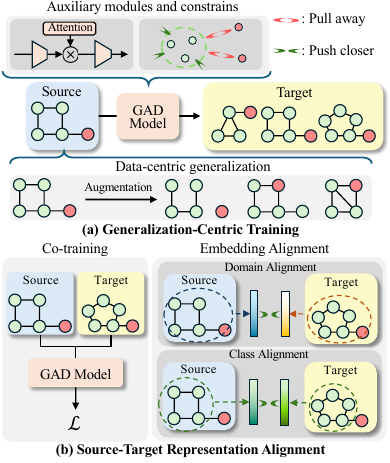}
\vspace{-3mm}
\caption{Sketch maps of transferable knowledge learning.}
\vspace{-3mm}
\label{fig:tk}
\end{figure}

\subsection{Transferable Knowledge Learning}
Learning transferable knowledge that generalizes to the target graph is central to effective transfer learning in GAD. Based on the availability of target data during training, existing methods fall into two sub-groups: \textit{generalization-centric training}, which focuses on enhancing the versatility of a pre-trained GAD model without access to target graphs; and \textit{source-target representation alignment}, which assumes target graphs are available to be aligned with the source ones. 

\begin{figure*}
\centering
\includegraphics[width=1\linewidth]{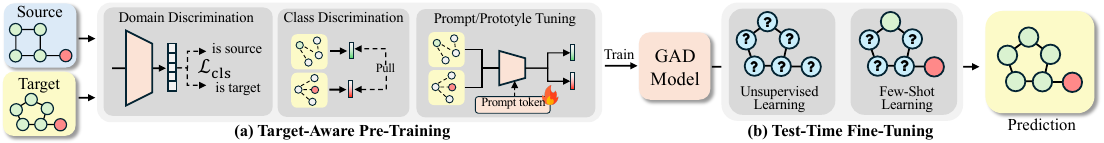}
\vspace{-4mm}
\caption{Sketch maps of target-specific patterns capturing.}
\vspace{-2mm}
\label{fig:ts}
\end{figure*}

\subsubsection{Generalization-Centric Training} %
To address the dynamic nature of real-world graph applications, several works focus on improving the generalizability of GAD models during the pre-training stage, without access to the target graph. 
The goal is to equip models with robust and transferable representations that can generalize to unseen graphs and anomalies as much as possible before any potential fine-tuning. 

A branch of methods enhances the generalization ability of the feature encoder by incorporating additional modules or training constraints. For instance, AdaGraph-T3~\cite{pirhayati2025graph} improves the graph encoder with a normal structure-preserved attention weighting module and class-aware regularization, which suppresses the influence of anomaly on normal features. 
Similarly, NSReg~\cite{wang2024nsreg} incorporates a normal structure regularization module to improve the generalizability against unseen anomaly types. 
Tailored by guarding graph application from out-of-distribution data, GOODAT~\cite{wang2024goodat} directly adapts a GNN model that has been well-trained for the application task and repurposes it for GAD. 

Instead of introducing additional modules to the architecture of graph encoders, another line of work improves generalizability from a data-centric perspective. By integrating data augmentation during training, these methods offer a more lightweight and plug-and-play approach to enhancing model robustness. 
AugAN~\cite{zhou2023improving} utilizes graph augmentation with episodic training, where discovered anomaly subgraphs are merged with augmented normal background subgraphs to enhance data diversity. 
In contrast, HGIF~\cite{ren2024heterophilic} specifically targets the heterophily shift between training and testing graphs in the context of graph fraud detection. To tackle this issue, edge-aware augmentation is applied to generate multiple virtual training environments with diverse levels of heterophily. 
Generalization-centric training makes it possible to enhance generalizability without access to target graphs during training. This preserves the data privacy of the target application and enables the pre-trained model to be deployed across a wide range of unseen graphs. However, these methods typically rely on heuristic assumptions about potential distribution shifts and graph properties, which may not align with the actual discrepancies encountered at test time. In practice, it is often feasible to overcome the challenge by obtaining a limited amount of target graph data, which can provide valuable insights into the actual target graph information. 

\subsubsection{Source-Target Representation Alignment} 
This subsection reviews the GAD methods designed for scenarios where the target graph is available during training, known as cross-domain GAD. In such cases, the key challenge lies in extracting transferable knowledge from additional training data, i.e., source graphs. To facilitate positive transfer, these methods typically use a shared encoder to align both the source and target graphs into the same representation space, followed by an auxiliary classification loss function to incorporate extra annotations.

One effective solution is to implicitly align features via self-supervised learning on both source and target graphs using a shared encoder. For instance, COMMANDER~\cite{ding2021cross} pioneers this approach by training the shared encoder using a feature reconstruction task on both source and target graphs, while ARMET~\cite{li2024cross} extends this methodology to graph-level anomaly detection by employing a one-class classification loss during pre-training. 
Another set of methods explicitly aligns features across domains. For example, CDFS-GAD~\cite{ChenCDFS} directly aligns overall graph representations across domains by utilizing an inter-domain graph contrastive learning loss, while ACT~\cite{wang2023cross} employs anomaly-aware one-class domain alignment to match the normal class between the two domains, allowing the model to generalize across diverse anomaly distributions.

\subsection{Target-Specific Patterns Capturing}
Apart from learning transferable knowledge, capturing target graph-specific patterns is also necessary for successful transfer learning, especially under limited data and annotations. Based on the availability of target graphs during pre-training, existing methods can be categorized into two subtypes: \textit{target-aware pre-training} that emphasizes learning target-specific patterns during cross-domain training, and \textit{test-time fine-tuning}, which adapts a pre-trained model to the target domain without access to pre-training data. 

\subsubsection{Target-Aware Pre-Training}
Target-aware pre-training aims to identify target-specific patterns by incorporating additional training objectives or modules during cross-domain training. These methods avoid the excessive alignment~\cite{xiao2021dynamic} and ensure the model does not overfit to the source domain but instead prioritizes improving GAD performance on the target graph. 

A significant portion of these methods employs auxiliary training tasks to address domain shift. For example, COMMANDER~\cite{ding2021cross} introduces an auxiliary domain discrimination task to guide the model in learning target-specific features within an adversarial training framework.
However, it may overlook semantic differences between anomalies and normal instances in the target domain, leading to less discriminative embeddings.
 To overcome this limitation, ARMET~\cite{li2024cross} aligns the centroids of normal graph embeddings across domains while separating those of anomalies. 
A similar class-distribution-centric idea is employed in the self-labeling-based deviation learning of ACT~\cite{wang2023cross}, which refines the learned prior knowledge of anomalies by focusing on nodes with high prediction confidence in each class, thereby generalizing the heuristics of their respective class distributions. While the aforementioned methods incorporate additional constraints to guide encoder training and improve GAD performance on the target graph, the limited expressiveness of the shared encoder can become a bottleneck, motivating researchers to enhance its capacity.

Inspired by the success of model repurposing techniques in NLP, recent works have begun to explicitly decouple class- or domain-specific knowledge to enhance the expressiveness of the encoder. For example, CDFS-GAD~\cite{ChenCDFS} adapts the idea of prompt learning by assigning a unique trainable prompt token to each domain. These tokens are used to compute attentional weights during message aggregation to enhance features with domain-specific patterns. Meanwhile, GDN~\cite{gao2023alleviating} decomposes class-specific knowledge from both feature and prototype perspectives to mitigate the heterophily shift problem in GAD. It disentangles node features into anomaly-relevant and irrelevant components to prevent contamination of normal embeddings, and iteratively computes class prototypes to enhance generalization under varying heterophily.

\subsubsection{Test-Time Fine-Tuning}

While the target-aware pre-training methods address limited annotations in the target graph, they rely on additional source-domain graphs, 
which reduces flexibility and limits applicability when source data is privacy-sensitive. 
To overcome these, test-time fine-tuning is another effective solution that adapts GAD models to the target graph using unsupervised or few-shot learning to reduce domain bias and improve GAD performance~\cite{zheng2025test,liu2025test,zheng2024online}.

As a representative approach, AdaGraph-T3~\cite{pirhayati2025graph} achieves representation adaptation by directly fine-tuning the encoder using a local affinity loss reweighted by the pseudo anomaly labels. 
Similarly, MetaGAD~\cite{xu2024metagad} also focuses on representation adaptation but few-shot annotations for training. It employs a meta learning framework to fine-tune a pre-trained GAD model by synergistically optimizing the anomaly detector and the meta learner. In contrast, GOODAT~\cite{wang2024goodat} repurposes the pre-trained GNN classifier without fine-tuning. It proposes an informative subgraph masking module trained to identify informative subgraphs using Graph Information Bottleneck-boosted losses. The anomaly score is estimated based on the loss, which measures the uncertainty in GNN predictions and the divergence between the original graph and the extracted subgraph embeddings.

\section{Foundation Models for GAD}

GAD foundation models aim to learn one-for-all models to achieve broad generalization. Unlike transfer learning that requires graphs from the same application scenario with aligned feature semantics and dimensions, they generalize across heterogeneous and unaligned patterns in diverse graph applications~\cite{qiao2025anomalygfm}. According to their generalization objective, existing methods in this category can be grouped into two subcategories: \textit{cross-granularity GAD foundation models}, which detects multiple levels of anomalies with a single framework, and \textit{cross-scenario GAD foundation models}, which learns one GAD model for datasets from various application domains.

\subsection{Cross-Granularity GAD Foundation Models}
\begin{figure}
\centering
\includegraphics[width=0.9\linewidth]{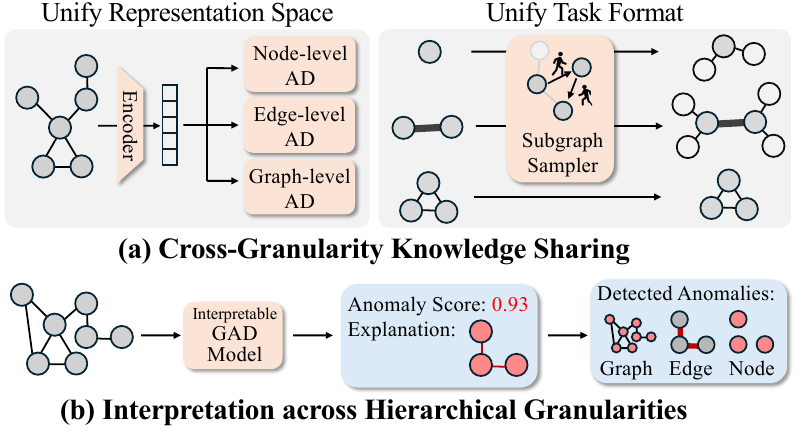}
\vspace{-3mm}
\caption{Sketch maps of cross-granularity GAD foundation models.}
\vspace{-3mm}
\label{fig:tax:uni}
\end{figure}

The cross-granularity GAD foundation models aim to integrate the detection of anomalies at different granularities, i.e., node-level, edge-level, and graph-level anomalies, into a single framework. Such a multi-task framework allows it to leverage the underlying correlations across different granularities, resulting in more accurate detection. Based on how these methods leverage the cross-granularity correlation, we categorize existing works into two sub-categories: \textit{enhancing GAD performance through shared knowledge} (Fig.~\ref{fig:tax:uni}(a)) and \textit{improving interpretability by considering hierarchical relationships} (Fig.~\ref{fig:tax:uni}(b)).

 \begin{figure*}

\includegraphics[width=1.0\linewidth]{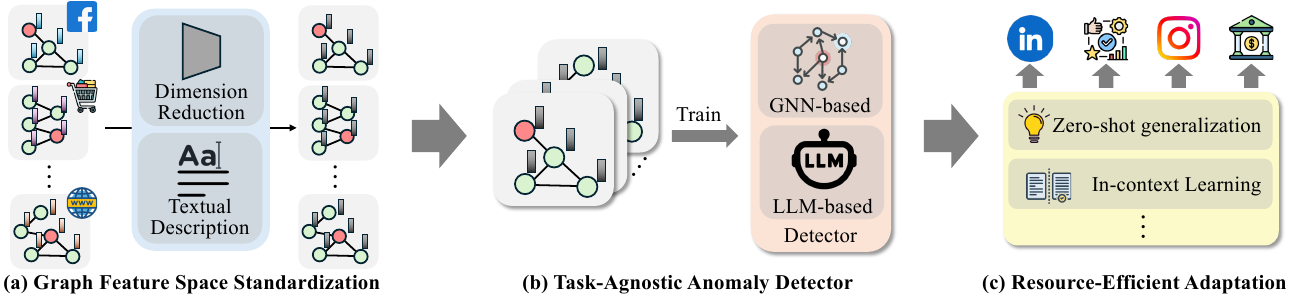}
\vspace{-5mm}
\caption{Sketch maps of cross-scenario GAD foundation models.}
\vspace{-3mm}
\label{fig:tax:FM}
\end{figure*}
\subsubsection{Cross-Granularity Knowledge Sharing}
Different levels of graph patterns often carry complementary information. For example, in a social network, an anomalous edge may signal a fraudster attempting to connect with others, while an anomalous community could comprise a cluster of such fraudulent accounts. Effectively leveraging such complementary information is the key to enhancing the performance of cross-granularity detection. 

Several works adopt a multi-task architecture, such as graph autoencoder (GAE), where a shared encoder is paired with separate detection heads for different granularities. This allows each head to specialize while preserving a common representation space across tasks. 
For example, HO-GAT~\cite{huang2021hybrid}  facilitates knowledge sharing across granularities through a hybrid-order attention mechanism that models node–motif interactions, which is trained with both multi-level reconstruction losses along with granularity-specific decoders for anomaly scoring. 
Similarly, HeagNet~\cite{fathony2024simultaneously} also adopts a GAE architecture but is tailored for detecting node- and edge-level in heterogeneous graphs. Differently, BOURNE~\cite{liu2024bourne} proposed a multi-view architecture that exploits complementary information across granularities via contrastive learning across views. It samples neighborhood subgraphs to build hypergraph views, where nodes represent edges from the original graph view. Anomaly scores are then computed by measuring embedding inconsistency between a sample and both its context and subgraph embedding in the opposite view, where the sample can be a node or an edge.

Motivated by the recent progress on graph foundational models, recent works unify the input task format across different anomaly granularities, enabling the encoder to handle all types consistently and improve cross-granularity generalization. For example, UniGAD~\cite{lin2024unigad} unifies multi-level anomaly detection by converting node- and edge-level tasks into graph-level ones through rooted subtree sampling. 
The sampler is optimized via the Rayleigh quotient to ensure that the resulting subtrees are informative for anomaly detection. 
Similarly, UniFORM~\cite{song2025uniform} unifies anomaly detection tasks into the graph level by constructing a subgraph pool using ego-neighbor graphs and random walk sampled subgraphs. 

\subsubsection{Anomaly Interpretation across Hierarchical 
Granularities}
As the famous saying goes, \textit{a small leak can sink a great ship.}  
In the context of GAD, high-level anomalies, such as graphs or subgraphs, are often inherently composed of low-level anomalies like nodes or edges. Leveraging this hierarchical relationship not only facilitates cross-granularity generalization but also offers valuable interpretability by explaining high-level anomalies through their low-level constituent parts. 

SIGNET~\cite{liu2023towards} pioneers this field by using the most informative subgraph to explain graph-level anomalies. It leverages contrastive learning between graph and hypergraph views to train a bottleneck subgraph extractor, enabling the simultaneous learning of node- and edge-level anomaly patterns. GRAM~\cite{yang2024gram} introduces a framework that integrates a node-level GAD model to provide fine-grained explanations for detected anomalous graphs. In contrast, ASD-HC~\cite{sun2024anomaly} builds up the maximum anomaly subgraph from node-level anomaly. It first detects node-level anomaly through a contrastive learning-based anomaly detector, and then uses these nodes as starting points for random walks to iteratively sample candidates for maximum anomaly subgraphs.

\subsection{Cross-Scenario GAD Foundation Models}
The cross-scenario GAD foundation models focus on detecting anomalies across different \textit{application scenarios} with a single model. These settings are challenging due to the lack of alignment in feature semantics and dimensions, as well as varying anomaly patterns. For example, in social networks, spamming behavior is easy to detect through unusually high posting rates, whereas financial fraud, such as money laundering, manifests in more subtle and complex patterns. To achieve effective generalization, GAD methods need to address three key challenges: \textit{unifying graph features across domains}, \textit{learning generalizable anomaly detectors}, and \textit{adapting to target datasets}.

\subsubsection{Graph Feature Space Standardization} 
Graph data from different domains often has diverse feature dimensions and semantics, making it challenging to standardize them within a common input space. Thus, many techniques have been proposed to standardize graph features across domains.

A common solution is applying dimensionality reduction algorithms, such as principal components analysis (PCA) or singular value decomposition (SVD), to unify the dimensionality of features across domains. For example, GUDI~\cite{li2024graph} selects the top feature dimensions with the highest variance to retain the most discriminative information for GAD. ARC~\cite{liu2024arc} and AnomalyGFM~\cite{qiao2025anomalygfm} further unify feature semantics across domains by leveraging graph homophily, with ARC computing feature-level smoothness and AnomalyGFM capturing the residual signal by subtracting the average neighbor feature. 
Inspired by batch normalization, UNPrompt~\cite{niu2024zero} rescales the transformed features using their mean and variance, calibrating semantic differences across domains.

With the help of pre-trained language encoders~\cite{pan2024integrating}, another line of work leverages text as a pivot to unify feature attributes across graphs. This is especially useful for GAD applications that naturally involve textual information. For example, GRACE~\cite{lu2024grace} detects software vulnerabilities using code property graphs containing meaningful textual descriptions associated with nodes and edges. When textual features are not available, they can be generated from graph attributes. For example, Wild-GAD~\cite{CaoWildGAD} describes node attributes in text using tabular headers as semantic cues.

\subsubsection{Task-Agnostic Anomaly Detector}
Even with a unified feature space, anomaly patterns often differ across domains, making it challenging to model them within a single framework. To address this,  generalized anomaly detectors aim to handle the distribution and pattern shifts across domains. Therefore, existing approaches aim to learn domain-invariant representations while capturing generalizable class semantics that distinguish anomalous from normal patterns.

Task-agnostic detectors often rely on carefully designed architectures and self-supervised objectives. 
For example, GUDI~\cite{li2024graph} models anomalies as information discarded during the denoising process of a diffusion-based graph autoencoder, effectively unifying features from diverse domains.
Similarly, UNPrompt~\cite{niu2024zero} adopts a graph contrastive learning-based anomaly detector~\cite{liu2021anomaly} that aligns augmented and original graph embeddings for unsupervised pre-training. When annotations are available, supervised objectives can further enhance generalizability. UNPrompt~\cite{niu2024zero} further utilizes trainable prompts to enhance the discriminability between normal and anomalous features among diverse domains, which are optimized by maximizing their similarity with the embeddings of the corresponding classes. Differently, AnomalyGFM~\cite{qiao2025anomalygfm} models generalizable class features using trainable prototypes, which are optimized to align with the residual features of annotated nodes.
ARC~\cite{liu2024arc} uses few-shot supervision as prototypes and optimizes with a marginal cosine similarity loss to encourage discriminative feature representations. 

Another potential solution is to leverage the multitask capabilities of large language models (LLMs), which inherently generalize across tasks and domains. For example, GRACE~\cite{lu2024grace} uses in-context learning for vulnerability detection. It directly integrates code snippets, code property graphs, and demonstrations into prompts, allowing the LLM to generate predictions without fine-tuning. Similarly, AnomalyLLM~\cite{liuAnomalyLLM} employs text prototype reprogramming to refine graph vocabulary and enhance alignment between graph and text modalities.

Last but not least, a few recent works move beyond the conventional training paradigm. For instance, TFGAD~\cite{zhou2025training}, directly uses the reconstruction error from SVD as the anomaly score, eliminating any training overhead. AD-Agent~\cite{yang2025ad} takes a meta approach by using LLMs to generate anomaly detection programs through multi-agent collaboration, leveraging their strengths in retrieval and code generation.
This method has demonstrated solid performance on PyGOD~\cite{liu2024pygod}, a widely recognized benchmark for GAD.

\subsubsection{Resource-Efficient Adaptation}
Supported by task-agnostic anomaly detectors, UNPrompt~\cite{niu2024zero} can achieve promising zero-shot detection performance on unseen graphs. Moreover, annotating a small amount of data in the target application can further improve GAD performance. Therefore, several methods explore training-free adaptation via in-context learning~\cite{liu2024arc, qiao2025anomalygfm, liuAnomalyLLM, lu2024grace}. Looking forward, it will be promising to see future research introduce more resource-efficient fine-tuning strategies into the GAD domain to further improve generalization.

\section{Challenges and Future Directions}
Generalization in GAD remains an evolving research frontier. Despite notable progress in both transfer learning and foundational models, many important challenges are still open and warrant further investigation in future research.

\noindent\textbf{Theoretical Guarantees.} 
While transfer learning and foundation models have shown empirical success in GAD, the theoretical understanding of transferability remains limited. It is still unclear why some auxiliary datasets yield positive transfer while others degrade performance. Factors such as domain-relatedness and anomaly semantics are believed to matter, yet formal definitions and theoretical justifications are still lacking. So far, only~\cite{CaoWildGAD} has explored this issue via heuristic data selection strategies. In this case, future research is encouraged to develop rigorous theories and principled methods for characterizing and improving transferability in GAD.

\noindent\textbf{Comprehensive Evaluation Protocols.} 
Despite the progress in improving GAD generalizability, standardized evaluation protocols remain lacking. 
This is largely due to the diversity in task settings and learning paradigms, which makes fair comparison across methods challenging and often ambiguous. Advancing the field requires unified protocols that capture various distribution shifts and generalization scenarios. While advanced studies~\cite{wang2025unifying,liu2022bond,tang2023gadbench} have benchmarked node-level and graph-level GAD, broader efforts are needed to encompass generalized application scenarios.

\noindent\textbf{Universal GAD Foundation Models.} 
While significant progress has been made in GAD foundation models, reaching the goal of ``one-for-all'' GAD models for all tasks and scenarios remains a key challenge. This goal can be achieved from two perspectives. From the model perspective, designing scalable and extendable architectures that obey scaling laws, as evidenced in other foundation model domains, represents a promising avenue to build more universal and capable GAD models.  
From the data perspective, training such models requires access to diverse and high-quality datasets that reflect a wide range of anomaly patterns in real-world graphs, highlighting the importance of comprehensive data collection for future research.

\noindent\textbf{Human-in-the-Loop (HITL).} 
Current generalized GAD models may struggle with complex, dynamic real-world data. To address this, integrating a HITL mechanism can enhance their robustness by providing real-time feedback through labeling and error correction. An early attempt~\cite{han2023anomaly} demonstrates this potential by incorporating human expertise into anomaly detection by investigating filtered anomalies and identifying shifts in normality. To build powerful GAD foundation models, HITL can be further explored and systematically incorporated in future studies.

\bibliographystyle{plain}
\bibliography{refs}

\begin{thebibliography}{10}

\bibitem{CaoWildGAD}
Yuxuan Cao, Jiarong Xu, Chen Zhao, Jiaan Wang, Carl Yang, Chunping Wang, and Yang Yang.
\newblock How to use graph data in the wild to help graph anomaly detection?
\newblock In {\em KDD}, page 61–72, 2025.

\bibitem{ChenCDFS}
Jiazhen Chen, Sichao Fu, Zhibin Zhang, Zheng Ma, Mingbin Feng, Tony~S. Wirjanto, and Qinmu Peng.
\newblock Towards cross-domain few-shot graph anomaly detection.
\newblock In {\em ICDM}, pages 51--60, 2024.

\bibitem{ding2024divide}
Kaize Ding, Xiaoxiao Ma, Yixin Liu, and Shirui Pan.
\newblock Divide and denoise: Empowering simple models for robust semi-supervised node classification against label noise.
\newblock In {\em KDD}, pages 574--584, 2024.

\bibitem{ding2021cross}
Kaize Ding, Kai Shu, Xuan Shan, Jundong Li, and Huan Liu.
\newblock Cross-domain graph anomaly detection.
\newblock {\em TNNLS}, 33(6):2406--2415, 2021.

\bibitem{fathony2024simultaneously}
Rizal Fathony, Jenn Ng, and Jia Chen.
\newblock Simultaneously detecting node and edge level anomalies on heterogeneous attributed graphs.
\newblock In {\em IJCNN}, pages 1--10. IEEE, 2024.

\bibitem{gao2023alleviating}
Yuan Gao, Xiang Wang, Xiangnan He, Zhenguang Liu, Huamin Feng, and Yongdong Zhang.
\newblock Alleviating structural distribution shift in graph anomaly detection.
\newblock In {\em WWW}, pages 357--365, 2023.

\bibitem{han2023anomaly}
Dongqi Han, Zhiliang Wang, Wenqi Chen, Kai Wang, Rui Yu, Su~Wang, Han Zhang, Zhihua Wang, Minghui Jin, Jiahai Yang, et~al.
\newblock Anomaly detection in the open world: Normality shift detection, explanation, and adaptation.
\newblock In {\em NDSS}, 2023.

\bibitem{huang2021hybrid}
Ling Huang, Ye~Zhu, Yuefang Gao, Tuo Liu, Chao Chang, Caixing Liu, Yong Tang, and Chang-Dong Wang.
\newblock Hybrid-order anomaly detection on attributed networks.
\newblock {\em TKDE}, 35(12):12249--12263, 2021.

\bibitem{li2025assemble}
Shiyuan Li, Yixin Liu, Qingsong Wen, Chengqi Zhang, and Shirui Pan.
\newblock Assemble your crew: Automatic multi-agent communication topology design via autoregressive graph generation.
\newblock {\em arXiv}, 2025.

\bibitem{li2024graph}
Xujia Li and Lei Chen.
\newblock Graph anomaly detection with domain-agnostic pre-training and few-shot adaptation.
\newblock In {\em ICDE}, pages 2667--2680. IEEE, 2024.

\bibitem{li2021relevance}
Yangyang Li, Yipeng Ji, Shaoning Li, Shulong He, Yinhao Cao, Yifeng Liu, Hong Liu, Xiong Li, Jun Shi, and Yangchao Yang.
\newblock Relevance-aware anomalous users detection in social network via graph neural network.
\newblock In {\em IJCNN}, pages 1--8. IEEE, 2021.

\bibitem{li2024cross}
Zhong Li, Sheng Liang, Jiayang Shi, and Matthijs van Leeuwen.
\newblock Cross-domain graph level anomaly detection.
\newblock {\em TKDE}, 2024.

\bibitem{lin2024unigad}
Yiqing Lin, Jianheng Tang, Chenyi Zi, H.~Vicky Zhao, Yuan Yao, and Jia Li.
\newblock Uni{GAD}: Unifying multi-level graph anomaly detection.
\newblock In {\em NeurIPS}, 2024.

\bibitem{liu2024bourne}
Jie Liu, Mengting He, Xuequn Shang, Jieming Shi, Bin Cui, and Hongzhi Yin.
\newblock Bourne: Bootstrapped self-supervised learning framework for unified graph anomaly detection.
\newblock In {\em ICDE}, pages 2820--2833. IEEE, 2024.

\bibitem{liu2024pygod}
Kay Liu, Yingtong Dou, Xueying Ding, Xiyang Hu, Ruitong Zhang, Hao Peng, Lichao Sun, and Philip~S Yu.
\newblock Pygod: A python library for graph outlier detection.
\newblock {\em JMLR}, 25(141):1--9, 2024.

\bibitem{liu2022bond}
Kay Liu, Yingtong Dou, Yue Zhao, Xueying Ding, Xiyang Hu, Ruitong Zhang, Kaize Ding, Canyu Chen, Hao Peng, Kai Shu, et~al.
\newblock Bond: Benchmarking unsupervised outlier node detection on static attributed graphs.
\newblock In {\em NeurIPS}, volume~35, pages 27021--27035, 2022.

\bibitem{liuAnomalyLLM}
Shuo Liu, Di~Yao, Lanting Fang, Zhetao Li, Wenbin Li, Kaiyu Feng, Xiaowen Ji, and Jingping Bi.
\newblock Anomalyllm: Few-shot anomaly edge detection for dynamic graphs using large language models.
\newblock In {\em ICDM}, pages 785--790, 2024.

\bibitem{liu2025test}
Yating Liu, Xin Zheng, Yi~Li, and Yanqing Guo.
\newblock Test-time adaptation on recommender system with data-centric graph transformation.
\newblock {\em IJCAI}, 2025.

\bibitem{liu2024self}
Yixin Liu, Thalaiyasingam Ajanthan, Hisham Husain, and Vu~Nguyen.
\newblock Self-supervision improves diffusion models for tabular data imputation.
\newblock In {\em CIKM}, pages 1513--1522, 2024.

\bibitem{liu2023towards}
Yixin Liu, Kaize Ding, Qinghua Lu, Fuyi Li, Leo~Yu Zhang, and Shirui Pan.
\newblock Towards self-interpretable graph-level anomaly detection.
\newblock In {\em NeurIPS}, volume~36, pages 8975--8987, 2023.

\bibitem{liu2024arc}
Yixin Liu, Shiyuan Li, Yu~Zheng, Qingfeng Chen, Chengqi Zhang, and Shirui Pan.
\newblock {ARC}: A generalist graph anomaly detector with in-context learning.
\newblock In {\em NeurIPS}, 2024.

\bibitem{liu2021anomaly}
Yixin Liu, Zhao Li, Shirui Pan, Chen Gong, Chuan Zhou, and George Karypis.
\newblock Anomaly detection on attributed networks via contrastive self-supervised learning.
\newblock {\em TNNLS}, 33(6):2378--2392, 2021.

\bibitem{liu2025graph}
Yixin Liu, Guibin Zhang, Kun Wang, Shiyuan Li, and Shirui Pan.
\newblock Graph-augmented large language model agents: Current progress and future prospects.
\newblock {\em arXiv}, 2025.

\bibitem{lu2024grace}
Guilong Lu, Xiaolin Ju, Xiang Chen, Wenlong Pei, and Zhilong Cai.
\newblock Grace: Empowering llm-based software vulnerability detection with graph structure and in-context learning.
\newblock {\em Journal of Systems and Software}, 212:112031, 2024.

\bibitem{ma2021comprehensive}
Xiaoxiao Ma, Jia Wu, Shan Xue, Jian Yang, Chuan Zhou, Quan~Z Sheng, Hui Xiong, and Leman Akoglu.
\newblock A comprehensive survey on graph anomaly detection with deep learning.
\newblock {\em TKDE}, 35(12):12012--12038, 2021.

\bibitem{miao2025blindguard}
Rui Miao, Yixin Liu, Yili Wang, Xu~Shen, Yue Tan, Yiwei Dai, Shirui Pan, and Xin Wang.
\newblock Blindguard: Safeguarding llm-based multi-agent systems under unknown attacks.
\newblock {\em arXiv}, 2025.

\bibitem{niu2024zero}
Chaoxi Niu, Hezhe Qiao, Changlu Chen, Ling Chen, and Guansong Pang.
\newblock Zero-shot generalist graph anomaly detection with unified neighborhood prompts.
\newblock {\em IJCAI}, 2025.

\bibitem{pan2025label}
Junjun Pan, Yixin Liu, Xin Zheng, Yizhen Zheng, Alan Wee-Chung Liew, Fuyi Li, and Shirui Pan.
\newblock A label-free heterophily-guided approach for unsupervised graph fraud detection.
\newblock In {\em AAAI}, volume~39, pages 12443--12451, 2025.

\bibitem{pan2024integrating}
Shirui Pan, Yizhen Zheng, and Yixin Liu.
\newblock Integrating graphs with large language models: Methods and prospects.
\newblock {\em IEEE Intelligent Systems}, 39(1):64--68, 2024.

\bibitem{pirhayati2025graph}
Delaram Pirhayati and Arlei Silva.
\newblock Graph anomaly detection via adaptive test-time representation learning across out-of-distribution domains.
\newblock {\em arXiv}, 2025.

\bibitem{qiao2025anomalygfm}
Hezhe Qiao, Chaoxi Niu, Ling Chen, and Guansong Pang.
\newblock Anomalygfm: Graph foundation model for zero/few-shot anomaly detection.
\newblock In {\em KDD}, 2025.

\bibitem{qiao2024deep}
Hezhe Qiao, Hanghang Tong, Bo~An, Irwin King, Charu Aggarwal, and Guansong Pang.
\newblock Deep graph anomaly detection: A survey and new perspectives.
\newblock {\em TKDE}, 37(9):5106--5126, 2025.

\bibitem{ren2024heterophilic}
Lingfei Ren, Ruimin Hu, Zheng Wang, Yilin Xiao, Dengshi Li, Junhang Wu, Yilong Zang, Jinzhang Hu, and Zijun Huang.
\newblock Heterophilic graph invariant learning for out-of-distribution of fraud detection.
\newblock In {\em MM}, pages 11032--11040, 2024.

\bibitem{shen2025understanding}
Xu~Shen, Yixin Liu, Yiwei Dai, Yili Wang, Rui Miao, Yue Tan, Shirui Pan, and Xin Wang.
\newblock Understanding the information propagation effects of communication topologies in llm-based multi-agent systems.
\newblock In {\em EMNLP}, 2025.

\bibitem{song2025uniform}
Chuancheng Song, Xixun Lin, Hanyang Shen, Yanmin Shang, and Yanan Cao.
\newblock Uniform: Towards unified framework for anomaly detection on graphs.
\newblock In {\em AAAI}, volume~39, pages 12559--12567, 2025.

\bibitem{sun2024anomaly}
Ying Sun, Wenjun Wang, Nannan Wu, and Chunlong Bao.
\newblock Anomaly subgraph detection through high-order sampling contrastive learning.
\newblock In {\em IJCAI}, pages 2362--2369, 2024.

\bibitem{tang2023gadbench}
Jianheng Tang, Fengrui Hua, Ziqi Gao, Peilin Zhao, and Jia Li.
\newblock Gadbench: Revisiting and benchmarking supervised graph anomaly detection.
\newblock In {\em NeurIPS}, volume~36, pages 29628--29653, 2023.

\bibitem{wang2024goodat}
Luzhi Wang, Dongxiao He, He~Zhang, Yixin Liu, Wenjie Wang, Shirui Pan, Di~Jin, and Tat-Seng Chua.
\newblock Goodat: towards test-time graph out-of-distribution detection.
\newblock In {\em AAAI}, volume~38, pages 15537--15545, 2024.

\bibitem{wang2023cross}
Qizhou Wang, Guansong Pang, Mahsa Salehi, Wray Buntine, and Christopher Leckie.
\newblock Cross-domain graph anomaly detection via anomaly-aware contrastive alignment.
\newblock In {\em AAAI}, volume~37, pages 4676--4684, 2023.

\bibitem{wang2024nsreg}
Qizhou Wang, Guansong Pang, Mahsa Salehi, Xiaokun Xia, and Christopher Leckie.
\newblock Open-set graph anomaly detection via normal structure regularisation.
\newblock In {\em ICLR}, 2025.

\bibitem{wang2025unifying}
Yili Wang, Yixin Liu, Xu~Shen, Chenyu Li, Rui Miao, Kaize Ding, Ying Wang, Shirui Pan, and Xin Wang.
\newblock Unifying unsupervised graph-level anomaly detection and out-of-distribution detection: A benchmark.
\newblock In {\em ICLR}, 2025.

\bibitem{wu2024graph}
Man Wu, Xin Zheng, Qin Zhang, Xiao Shen, Xiong Luo, Xingquan Zhu, and Shirui Pan.
\newblock Graph learning under distribution shifts: A comprehensive survey on domain adaptation, out-of-distribution, and continual learning.
\newblock {\em arXiv}, 2024.

\bibitem{xiao2021dynamic}
Ni~Xiao and Lei Zhang.
\newblock Dynamic weighted learning for unsupervised domain adaptation.
\newblock In {\em CVPR}, pages 15242--15251, 2021.

\bibitem{xu2019relation}
Fengli Xu, Jianxun Lian, Zhenyu Han, Yong Li, Yujian Xu, and Xing Xie.
\newblock Relation-aware graph convolutional networks for agent-initiated social e-commerce recommendation.
\newblock In {\em CIKM}, pages 529--538, 2019.

\bibitem{xu2024metagad}
Xiongxiao Xu, Kaize Ding, Canyu Chen, and Kai Shu.
\newblock Metagad: Meta representation adaptation for few-shot graph anomaly detection.
\newblock In {\em DSAA}, pages 1--10. IEEE, 2024.

\bibitem{yang2025ad}
Tiankai Yang, Junjun Liu, Wingchun Siu, Jiahang Wang, Zhuangzhuang Qian, Chanjuan Song, Cheng Cheng, Xiyang Hu, and Yue Zhao.
\newblock Ad-agent: A multi-agent framework for end-to-end anomaly detection.
\newblock {\em arXiv}, 2025.

\bibitem{yang2024gram}
Yifei Yang, Peng Wang, Xiaofan He, and Dongmian Zou.
\newblock Gram: An interpretable approach for graph anomaly detection using gradient attention maps.
\newblock {\em Neural Networks}, 178, 2024.

\bibitem{zhao2025freegad}
Yunfeng Zhao, Yixin Liu, Shiyuan Li, Qingfeng Chen, Yu~Zheng, and Shirui Pan.
\newblock Freegad: A training-free yet effective approach for graph anomaly detection.
\newblock In {\em CIKM}, 2025.

\bibitem{zheng2025test}
Xin Zheng, Wei Huang, Chuan Zhou, Ming Li, and Shirui Pan.
\newblock Test-time graph neural dataset search with generative projection.
\newblock In {\em ICML}, 2025.

\bibitem{zheng2024online}
Xin Zheng, Dongjin Song, Qingsong Wen, Bo~Du, and Shirui Pan.
\newblock Online gnn evaluation under test-time graph distribution shifts.
\newblock In {\em ICLR}, 2024.

\bibitem{zhou2025training}
Cheng Zhou, Guangxia Li, Hao Weng, and Yiyu Xiang.
\newblock Training-free graph anomaly detection: A simple approach via singular value decomposition.
\newblock In {\em WWW}, pages 4196--4205, 2025.

\bibitem{zhou2022domain}
Kaiyang Zhou, Ziwei Liu, Yu~Qiao, Tao Xiang, and Chen~Change Loy.
\newblock Domain generalization: A survey.
\newblock {\em TPAMI}, 45(4):4396--4415, 2022.

\bibitem{zhou2023improving}
Shuang Zhou, Xiao Huang, Ninghao Liu, Huachi Zhou, Fu-Lai Chung, and Long-Kai Huang.
\newblock Improving generalizability of graph anomaly detection models via data augmentation.
\newblock {\em TKDE}, 35(12):12721--12735, 2023.

\end{thebibliography}

\end{document}